\documentclass[conference]{IEEEtran}
\pdfoutput=1    % For arXiv issues

\usepackage{xcolor}
\usepackage{subfig}
\usepackage{multirow}
\usepackage{multicol}
\definecolor{wong-black}        {HTML}{000000}
\definecolor{wong-lightorange}  {HTML}{E69F00}
\definecolor{wong-lightblue}    {HTML}{56B4E9}
\definecolor{wong-green}        {HTML}{009E73}
\definecolor{wong-yellow}       {HTML}{F0E442}
\definecolor{wong-darkblue}     {HTML}{0072B2}
\definecolor{wong-darkorange}   {HTML}{D55E00}
\definecolor{wong-pink}         {HTML}{CC79A7}

\usepackage[accsupp]{axessibility}  % Improves PDF readability for those with disabilities.

\usepackage{url}
\usepackage{hyperref} % Working hyperlink (https://www.overleaf.com/learn/latex/Hyperlinks)

\hypersetup{
    colorlinks=true,
    citecolor=wong-green,
    linkcolor=wong-darkblue,
    filecolor=wong-pink,      
    urlcolor=wong-pink
    }

\usepackage{cite}
\usepackage{amsmath,amssymb,amsfonts}
\usepackage{algorithmic}
\usepackage{graphicx}
\usepackage{booktabs}
\usepackage{textcomp}
\usepackage[nolist, nohyperlinks, printonlyused]{acronym} % For consistent acronyms
\usepackage{geometry}

\usepackage{pifont}% http://ctan.org/pkg/pifont
\newcommand{\cmark}{\ding{51}}%

\def\BibTeX{{\rm B\kern-.05em{\sc i\kern-.025em b}\kern-.08em
    T\kern-.1667em\lower.7ex\hbox{E}\kern-.125emX}}

\begin{document}
\bstctlcite{IEEEexample:BSTcontrol}
% -------------------------- TITLE -------------------------------

\newgeometry{left = 17.5mm, top=25.4mm, bottom=19.1mm, right=17.5mm}     % use whatever margins you want for left, right, top and bottom.

\title{On The Impact of Replacing Private Cars with Autonomous Shuttles: An Agent-Based Approach}

%\restoregeometry     %so it does not affect the rest of the pages.

% -------------------------- AUTHORS -------------------------------

\author{\IEEEauthorblockN{Daniel Bogdoll\IEEEauthorrefmark{2}\IEEEauthorrefmark{3}\textsuperscript{\textasteriskcentered},
Louis Karsch\IEEEauthorrefmark{3}\textsuperscript{\textasteriskcentered},
Jennifer Amritzer\IEEEauthorrefmark{2},
and J. Marius Zöllner\IEEEauthorrefmark{2}\IEEEauthorrefmark{3}}

\IEEEauthorblockA{\IEEEauthorrefmark{2}FZI Research Center for Information Technology, Germany\\
bogdoll@fzi.de}
\IEEEauthorblockA{\IEEEauthorrefmark{3}Karlsruhe Institute of Technology, Germany\\}}

\maketitle

\def\thefootnote{\textsuperscript{\textasteriskcentered}}\footnotetext{These authors contributed equally}\def\thefootnote{\arabic{footnote}}

% \nnfootnote{\textasteriskcentered~These authors contributed equally}

% -------------------------- ACRONYMS -------------------------------

\begin{acronym}
    \acro{ml}[ML]{Machine Learning}
	\acro{cnn}[CNN]{Convolutional Neural Network}
	\acro{dl}[DL]{Deep Learning}
	\acro{ad}[AD]{Autonomous Driving}
\end{acronym}

% -------------------------- ABSTRACT -------------------------------

\begin{abstract}
The European Green Deal aims to achieve climate neutrality by 2050, which demands improved emissions efficiency from the transportation industry. This study uses an agent-based simulation to analyze the sustainability impacts of shared autonomous shuttles. We forecast travel demands for 2050 and simulate regulatory interventions in the form of replacing private cars with a fleet of shared autonomous shuttles in specific areas. We derive driving-related emissions, energy consumption, and non-driving-related emissions to calculate life-cycle emissions. We observe reduced life-cycle emissions from 0.4\% to 9.6\% and reduced energy consumption from 1.5\% to 12.2\%.
\end{abstract}

% -------------------------- KEYWORDS -------------------------------


% -------------------------- CONTENT -------------------------------

\section{Introduction}
\label{sec:introduction}

While some research on the adoption of autonomous vehicles~(AV) predicts harmful environmental effects due to increased power usage or longer distances traveled~\cite{sudhakar2022data}, others believe that AVs can play a crucial role in lowering overall emissions~\cite{gawron2018life, massar2021impacts, patella2019carbon}. By enhancing driving efficiency, decreasing the necessity for cars, and improving traffic flow, positive outcomes can be achieved~\cite{silva2022environmental}. Since these benefits largely rely on the extent of AV adoption, our work examines the environmental effects of Shared Autonomous Vehicles~(SAV) by exploring different deployment scenarios through an agent-based simulation. In particular, we consider hypothetical policy regulations that prohibit private vehicles in both urban and rural areas and only allow the use of SAVs.

\textbf{Research Gap.}
Several related works analyze the effect of SAVs on traffic~\cite{bischoff2016autonomous, horl2017agent,liu2017tracking, zhang2015performance}. However, models are often built on outdated data~\cite{hidaka2018forecasting, lecureux2021sensitivity, vosooghi2019shared, ziemke2023accessibilities, zwick2021agent}, while it has been shown that travel demand forecasts can lead to more realistic scenario analyses~\cite{bansal2017forecasting, cokyasar2020analyzing, hao2017analysis, hidaka2018forecasting}. To steer AV adoption, political regulations are posed as a viable solution~\cite{geistfeld2018regulatory, hansson2020regulatory}. However, variations of scenarios are missing to derive an improved understanding of a wider range of possible SAV scenarios. Another line of work examines the effects on individual sustainability metrics~\cite{gruvzauskas2018minimizing, heard2018sustainability, williams2020assessing}. Holistic approaches cannot be found, as Table~\ref{tab:sota} shows.  

\textbf{Contribution.}
As we have shown, there are multiple challenges present in the state of the art. First, we propose a travel demand forecast for 2050 to build our simulation upon realistic data. Then, we perform a multitude of experiments based on an agent-based simulation, where we examine three different regulatory interventions. These describe SAV-exclusive traffic zones: The inner city of Berlin, the city of Berlin, and the metropolitan area of Berlin and Brandenburg, including both urban and rural traffic. Finally, we derive, among others, holistic sustainability implications. For more details, we refer to~\cite{Karsch_Sustainability_2023_MA}. All code can be found on \href{https://github.com/daniel-bogdoll/agent_based_av}{GitHub}

\begin{figure}[tb]
    
    \begin{center}
        \includegraphics[width=1.0\columnwidth,angle=0]{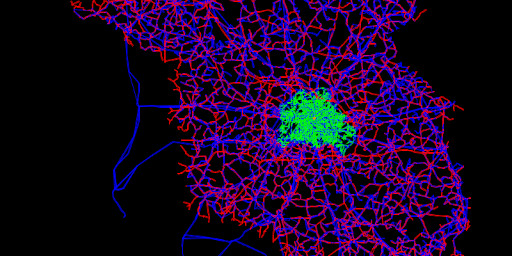}
        \caption{One of our simulated deployment strategies for shared autonomous shuttles. Here, private cars are banned in the city of Berlin, as indicated by the green streets. Red lines indicate streets for private cars and blue lines public transport routes. All trips need to be done with shared autonomous shuttles, bicycles, public transportation, or on foot. Adapted from~\cite{Karsch_Sustainability_2023_MA}.}
        \label{fig:berlin_city_area}
    \end{center}
    
\end{figure}

\section{Related Work}
\label{sec:related_work}

Our work lies at the intersection of travel demand forecasts, agent-based simulations of autonomous vehicles, and their sustainability impacts. In the following, we show related works in these fields and our contributions in Table~\ref{tab:sota}.

\begin{table*}[]
\resizebox{\textwidth}{!}{
    \begin{tabular}{llcccccccc}
    \toprule
        ~ & Year & Forecast  & \multicolumn{4}{c}{Agent-Based  Simulation}  & \multicolumn{3}{c}{Sustainability Impacts} \\ 
        \midrule
        ~ & ~ & ~ & \# SAV Zones & \# Scenarios & Urban & Rural & Energy & Tailpipe & Life-Cycle \\ \midrule 
        Cokyasar et al.~\cite{cokyasar2020analyzing} & 2020 & 2040 & -- & -- & -- & -- & \cmark & \cmark & --\\ 
        Hidaka and Shiga~\cite{hidaka2018forecasting} & 2018 & 2040 & -- & -- & -- & -- & -- & -- & -- \\
        Bridgelall and Stubbing.~\cite{bridgelall2021forecasting} & 2021 & 2050 & -- & -- & -- & -- & -- & -- & --    \\ 
        Lécureux and Kaddoura ~\cite{lecureux2021sensitivity} & 2021 & -- & 1 & 12 & \cmark & \cmark & -- & -- & --  \\ 
        Vosooghi et al.~\cite{vosooghi2019shared} & 2019 & -- & 1 & 37 & \cmark & \cmark & -- & -- & --  \\ 
        Ziemke and Bischoff~\cite{ziemke2023accessibilities} & 2023 & -- & 1 & 3 & \cmark & -- & -- & -- & --  \\ 
        Zwick et al.~\cite{zwick2021agent} & 2021 & -- & 2 & 5 & \cmark & -- & -- & -- & --    \\ 
        Liu et al.~\cite{liu2017tracking} & 2017 & -- & 1 & 4 & \cmark & \cmark &  \cmark & \cmark & --  \\ 
        Bischoff and Maciejewski~\cite{bischoff2016autonomous} & 2016 & -- & 2 & 3 & \cmark & -- & -- & -- & --  \\ 
        Hörl~\cite{horl2017agent} & 2017 & -- & 1 & 3 & \cmark & -- & -- & -- & --  \\ 
        Ben-Dor et al.~\cite{ben2022modal} & 2022 & -- & 1 & 8 & \cmark & \cmark & -- & -- & --  \\ 
        Kaddoura and Schlenther~\cite{kaddoura2021impact} & 2021 & -- & 1 & 15 & \cmark & \cmark & -- & -- & --  \\ 
        Li et al.~\cite{li2023simulation} & 2023 & -- & 1 & 8 & \cmark & -- & -- & -- & --  \\  
        Harper et al.~\cite{harper2018exploring} & 2018 & -- & 1 & 5 & \cmark & -- & \cmark   & \cmark   & --    \\ 
        Patella et al.~\cite{patella2019carbon} & 2019 & -- & 1 & 2 &  \cmark & -- & \cmark   & \cmark   & \cmark    \\ 
        Martinez and Viegas~\cite{martinez2017assessing} & 2017 & -- & 1 & 2 & \cmark & -- & -- & \cmark   & --  \\ \midrule
        Ours & 2023 & 2050 & 3 & 12 & \cmark & \cmark & \cmark & \cmark & \cmark \\ \bottomrule

\end{tabular}%
}
\caption{Related work with a focus on travel demand forecasts, agent-based simulations for AVs, and sustainability impacts.}
\label{tab:sota}
\end{table*}

\textbf{Travel Demand Forecast.} Often, a timeframe beyond 2030 is considered for AVs~\cite{bansal2017forecasting, cokyasar2020analyzing, litman2017autonomous}. Cokyasar et al.~\cite{cokyasar2020analyzing} examined AVs' energy consumption in 2025 and 2040 while factoring in changes in overall travel demand. Their analysis considered population growth, shifts in vehicle types, and advancements in technology as key factors influencing travel demand. Hidaka and Shiga~\cite{hidaka2018forecasting} consider factors such as shifts in age demographics, reduced travel needs due to aging, rising rates of driver's license surrender, and regional influences up until 2040. Bansal and Kockelman~\cite{bansal2017forecasting} projected potential shares of AVs in the private light-duty transport sector of the United States, considering aspects such as pricing and willingness to pay. 

However, these analyses are not agent-based. Microscopic simulations commonly build on outdated inputs, not taking into account any projections into the future~\cite{lecureux2021sensitivity,vosooghi2019shared, ziemke2023accessibilities, zwick2021agent}. 

\textbf{Agent-based (S)AV Simulation.}
In their analysis based on the Multi-Agent Transport Simulation~(MATSim), Liu et al.~\cite{liu2017tracking} examined the effect of different pricing strategies on SAV travel in Austin, Texas, ranging from 0.5~USD to 1.25~USD per mile. Based on their simulation, they see adoption ranges between 9.2\% and 50.9\%~\cite{liu2017tracking}. Ziemke and Bischoff~\cite{ziemke2023accessibilities} examine the effects of different SAV regulatory scenarios on accessibility in the MATSim Open Berlin Scenario~\cite{ziemke2019matsim}. They utilize the Demand-Responsive Transport (DRT) extension of MATSim, assuming single-passenger SAV service~\cite{ziemke2023accessibilities}. For instance, they implement regulatory interventions where a random selection of 10\% of agents that were using publication transportation are now being served by SAVs~\cite{ziemke2023accessibilities}, which does not take a holistic view but is an interesting approach. Using a Munich traffic simulation in MATSim, Zwick et al.~\cite{zwick2021agent} examined how SAVs affect traffic noise levels on a city-wide level~\cite{zwick2021agent}. The authors introduce SAVs by replacing all car plans of agents with SAV travel or giving free mode choice~\cite{zwick2021agent}. Similarly, Bischoff and Maciejewski~\cite{bischoff2016autonomous} conducted their SAV analysis using the MATSim DRT module. They simulate a scenario where all trips in the city of Berlin based on private cars are being replaced by an SAV fleet~\cite{bischoff2016autonomous}. 

In general, we observe a small variety of examined SAV introduction concepts in agent-based simulation, which do not allow for appropriate analyses of potential SAV scenarios and variations that come with it.

\textbf{Sustainability of AV and SAV travel.} For the analysis of sustainability impacts, agent-based simulations are very common~\cite{harper2018exploring, fagnant2014travel, patella2019carbon}. Harper et al. used traveled vehicle kilometers to calculate emissions and energy consumption for AVs. Their study found that energy use and greenhouse gas emissions would increase by up to 3\%~\cite{harper2018exploring}. The study considered solely driving-related emissions. Patella et al.~\cite{patella2019carbon} performed a simulation of traffic in Rome, including the life cycle impact of AVs. The study focused on quantifying car-related emission generation and energy consumption ~\cite{patella2019carbon}. They examined the impact of AVs, including a fully electrified fleet, in their experiments. However, the study only examined a hypothetical 100\% electric AV scenario.

\textbf{Research gaps.} As AV models often build upon outdated travel data~\cite{hidaka2018forecasting}, we conclude limited realism of SAV analyses. AV and SAV diffusion is assumed with unforeseen dynamics which is why, policies are posed to be a lever to model SAV introduction~\cite{bansal2017forecasting}. We identify the need for more SAV strategies using zone-based adoption steering. Furthermore, we recognize the need for research on more comprehensive sustainability analyses.
\section{Method}
\label{sec:method}

Addressing the need to incorporate future travel demand when analyzing the impact of AVs, we conduct a travel demand forecast until 2050. Additionally, we aim to narrow the gap of missing heterogeneity of various zone-based SAV scenarios by implementing four scales of SAV adoptions steered by non-SAV ban zones. Lastly, we work towards more holistic sustainability analyses by deriving sustainability impacts based on driving and non-driving related emissions as well as driving-related energy consumption. Instead of predicting adoption rates of AVs~\cite{hidaka2018forecasting,bansal2017forecasting}, we simulate political regulations that demand the deployment of autonomous shuttle fleets. Our approach is inspired by low-emission zones, which only allow access to certain urban areas in Germany for vehicles that fulfill emission-related criteria. In our simulations, we define city areas in which private passenger cars are banned~\cite{cruisetech}. Traversing these areas becomes only possible with shared autonomous shuttles, public transport, bicycles, and on foot. We propose a set of three scenarios for such areas: The inner city of Berlin, the city of Berlin, see Figure~\ref{fig:berlin_city_area}, and the metropolitan region of Berlin, including parts of Brandenburg. We benchmark these three scenarios against a case with no bans or shared autonomous vehicles. As it will take several more years before autonomous vehicles are available at scale in large cities, we also derive two forecasts for travel demand in 2050 based on different growth assumptions. We use an original 2011 scenario from Ziemke et al.~\cite{ziemke2019matsim} as a benchmark.

\begin{table}[t]
\resizebox{\columnwidth}{!}{%
\begin{tabular}{@{}llllr@{}}
\toprule
                       & \multicolumn{4}{c}{SAV Deployment Area}                                                            \\ \midrule
Travel Demand & \textbf{0: --} & \textbf{1: Center} & \textbf{2: City} & \textbf{3: Metro} \\
\textbf{1: 2011-Base}  & 0            & 27,300                     & 46,700              & 68,100                             \\
\textbf{2: 2050-Low}   & 0            & 29,000                     & 48,900              & 71,400                             \\
\textbf{3: 2050-High}  & 0            & 31,000                     & 52,300              & 76,800                             \\ \bottomrule
\end{tabular}%
}
\caption{Rounded numbers of autonomous shuttles deployed for our twelve scenarios. We examine three different travel demands, namely the original 2011 benchmark and both a conservative and a more optimistic forecast. For each, we analyze a benchmark scenario without SAVs and three different SAV deployment strategies within Berlin and Brandenburg.}
\label{tab:Experiments}
\end{table}

Subsequently, we perform an agent-based simulation of these $4\times3=12$ scenarios and compute their sustainability implications. Since agents in our simulated regulations require autonomous shuttles for longer distances, we deploy as many shuttles in our simulations as needed. However, in Section~\ref{sec:agentbased} we also examine the impact of reduced fleet sizes. Table~\ref{tab:Experiments} shows the number of deployed autonomous shuttles per scenario and introduces our naming scheme. In the following, we will refer to these twelve scenarios by SCX.Y, where SC1.0 describes the benchmark scenario with a travel demand from 2011 and no SAVs, while SC3.3 describes a high travel demand forecast for 2050 and the whole metropolitan region as the politically regulated zone.

\section{Travel Demand Forecast}
\label{sec:forecast}

\begin{figure}[htb]
    
    \begin{center}
        \includegraphics[width=1.0\columnwidth,angle=0]{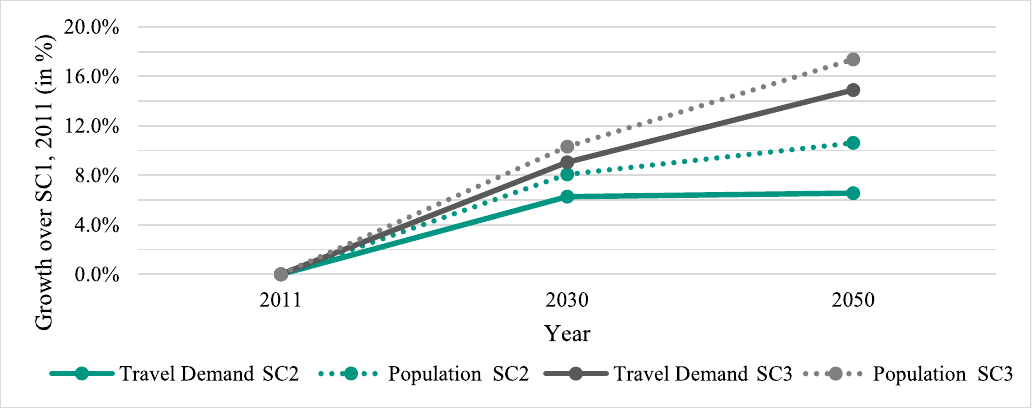}
        \caption{Conservative and optimistic travel demand forecasts of population growth (dotted lines) and travel demand (solid lines) for 2030 and 2050. Adapted from~\cite{Karsch_Sustainability_2023_MA}.}
        \label{fig:Travel demand and Population Growth}
    \end{center}
    
\end{figure}

While an aging population contributes to population growth, not all age groups have the same mobility needs. Thus, we project travel demand by combining population growth and age-related travel behavior. Due to the high uncertainty surrounding other impacts, such as vehicle type changes and technology improvements, we do not consider their influences. Instead, we formulate three population growth scenarios using models from the German Institute for Statistics~(Destatis)~\cite{genesis2023population}. By accounting for travel demand per age group, based on the German mobility report by infas and DLR~\cite{infas2017mobility}, we reduced the impact of population growth per age group with low average travel demand, e.g., the elderly population, and vice-versa. Therefore, we model more accurately future travel demand based on initial population growth. Figure~\ref{fig:Travel demand and Population Growth} shows the results of our travel demand forecast, where we also compute an intermediate forecast for 2030. However, for our simulated scenarios, we focus on the predictions for 2050.

Our optimistic forecast scenario shows an initial population growth of 17.4\% compared to 2011. Accounting for age-related travel demands results in a reduced increase of 14.9\%, which is the base for all SC3.Y scenarios. The group with the largest influence on this slower growth is the age group 40-49, as it makes up the largest part of the population and has the highest age-based travel demand. Our conservative forecast shows a much lower increase in travel demand where migration and emigration ratios are core differences. Since precise forecasts are not possible due to many parameters and high uncertainties, the consideration of these two different future scenarios enables us to depict developments within which the actual development of mobility demand will most likely take place.

\section{Agent-Based Simulation}
\label{sec:agentbased}

To simulate the mobility behavior of individual agents, we base our work on the MATSim~\cite{w2016multi} framework, which has been widely adopted for agent-based AV simulations~\cite{li2021systematic, jing2020agent}. Its ability to simulate complex networks and numerous agents makes it an ideal choice for AV analyses~\cite{li2021systematic}. We build upon the 10\% Open Berlin Scenario~\cite{ziemke2019matsim} with two iterations per scenario. We integrate an SAV fleet into the scenario using the DRT module. It allows for on-demand vehicle dispatching, making it particularly suitable for SAVs~\cite{li2021systematic}. The baseline scenario comprises 491,000 agents out of a total population of 5.7 million in Berlin and Brandenburg~\cite{einheit2015statistische}.

Given our computed travel demand forecasts from Section~\ref{sec:forecast}, we incorporate these macroscopic scenarios into the microscopic OpenBerlin baseline scenario from 2011. We achieve this by duplicating agents in our simulation to account for the travel demand increase. This leads to a 6.6\% increase for SC2.0-SC2.3 with a total of 523,763 agents and a 14.9\% increase for SC3.0-SC3.3 with a total of 564,513 agents. The additional home locations and departure times are randomly sampled from the existing agent home locations, resulting in a population density increase of 6.6\% and 14.9\%.

By comparing the two scenarios for 2050 without introducing autonomous shuttles with our benchmark SC1.0, we can determine the changed load on the transport infrastructure caused by the increased population. As detailed in~\cite{Karsch_Sustainability_2023_MA}, we see an increase in traffic of up to 12.4\% for SC2.0 and 22.7\% for SC3.0 for hourly traffic across all links, e.g., streets or public transport lines, until 8:00 pm. Especially for SC3.0, the infrastructure is reaching its limits, as after-hours traffic causes congestion that lasts well past midnight. Such a case requires more intelligent mobility solutions, such as the ones examined in our work.

As we have introduced our new populations, we present how we introduced the shuttles, see Table~\ref{tab:Experiments}, with DRT. For technical reasons, we allowed for a maximum fleet size of 100,000 SAVs, well above the maximum number of necessary vehicles in any scenario as computed by DRT. Agents can call for specific rides while the shuttle that creates the least workload will be chosen to execute the request. SAVs always stay at the last location of drop off until called again. Alternatively, MATSim can provide hubs for shuttles to return to. However, our experiments showed that this leads to high waiting times and additional vehicle kilometers, which is why we decided to use the default setting. We have analyzed both shuttles with four and six seats. Given the determined fleet sizes by the DRT module, only very few rides showed more than four passengers at a time. Thus, we performed our experiments with shuttles carrying up to four passengers.

\begin{figure}[t]
  \centering
  \subfloat[SC1.1.]{%
    \includegraphics[width=0.3\columnwidth]{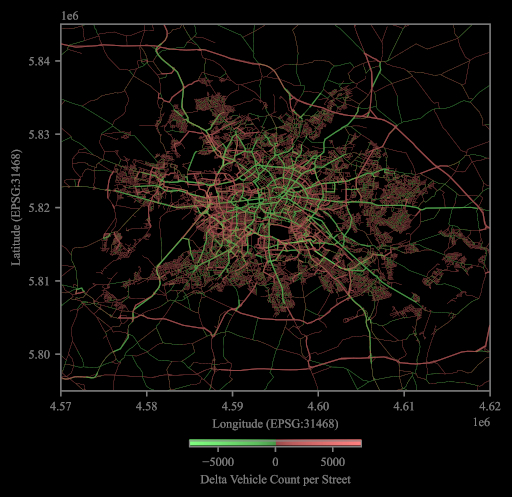}%
    \label{fig:Travel_Increase_SC1.1}
  }
  \hfill
  \subfloat[SC1.2.]{%
    \includegraphics[width=0.3\columnwidth]{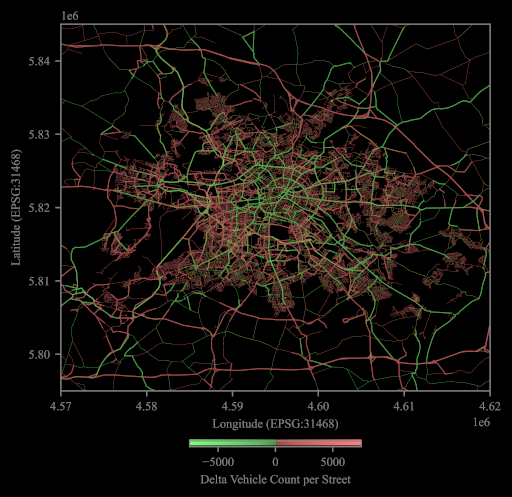}%
    \label{fig:Travel_Increase_SC1.2}
  }
  \hfill
  \subfloat[SC1.3.]{%
    \includegraphics[width=0.3\columnwidth]{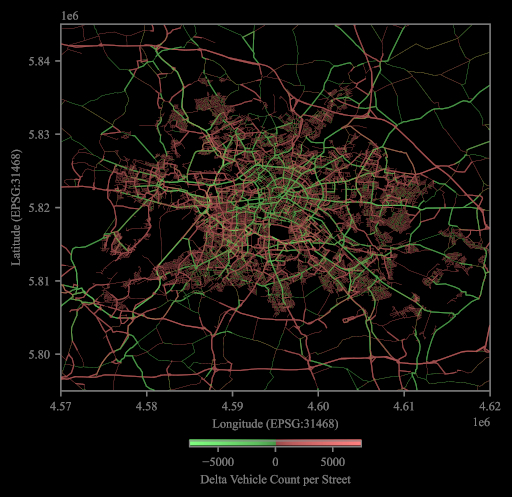}%
    \label{fig:Travel_Increase_SC1.3}
  }
  \caption{Delta of car-based travel with SAV introduction compared to the original OpenBerlin scenario, where green shows a reduced vehicle count per street and red an increased count. Brighter colors indicate higher values. Best viewed at 600\%.}
  \label{fig:Heatmap_Traffic_SAV_Level}
\end{figure}

Introducing shared autonomous shuttles in our agent-based simulation allows us to understand the system's dynamics on a very detailed level. First, we examine how SAVs contribute to changed usages of the available traffic infrastructure on the example of SC1.1. As visible in Figure~\ref{fig:Heatmap_Traffic_SAV_Level}, SAVs show increased usage of main roads around Berlin. On the other hand, main road usage within the city center is reduced compared to the original scenario.

\begin{figure}[t]
    
    \begin{center}
        \includegraphics[width=1.0\columnwidth,angle=0]{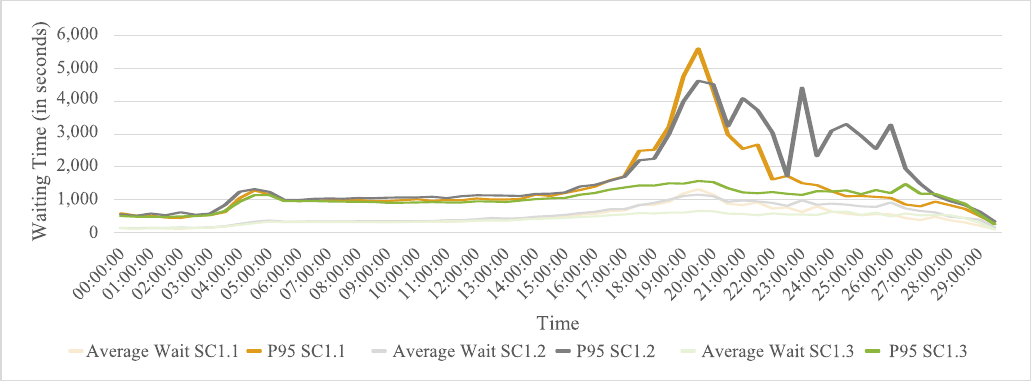}
        \caption{Waiting times per SAV scenario. Adapted from~\cite{Karsch_Sustainability_2023_MA}.}
        \label{fig:Evaluation_Waiting_Times}
    \end{center}
    
\end{figure}

After introducing the required fleets of shared autonomous shuttles, we examine their service levels for the calibrated 2011 scenario by analyzing how long agents wait for their shuttle to arrive after they have called it, similar to Bischoff and Maciejewski~\cite{bischoff2016autonomous}. As shown in Figure~\ref{fig:Evaluation_Waiting_Times}, we see high waiting times for SC1.1 and SC1.2. As vehicles stay close to their drop-off location at night, only minor waiting times can be seen in the morning. SC1.1 shows average waiting times of 22 minutes and up to 93 minutes for the 95th percentile. This can be explained by the lack of SAVs close by. For SC1.2 we observe constant high waiting times in the evening, as SAVs have to travel long distances. Only in SC1.3, where no more private cars are allowed, the fleet leads to significant improvements without any peaks.  

\begin{figure}[h]
  \centering
  \subfloat[SC3.1.]{%
    \includegraphics[width=0.3\columnwidth]{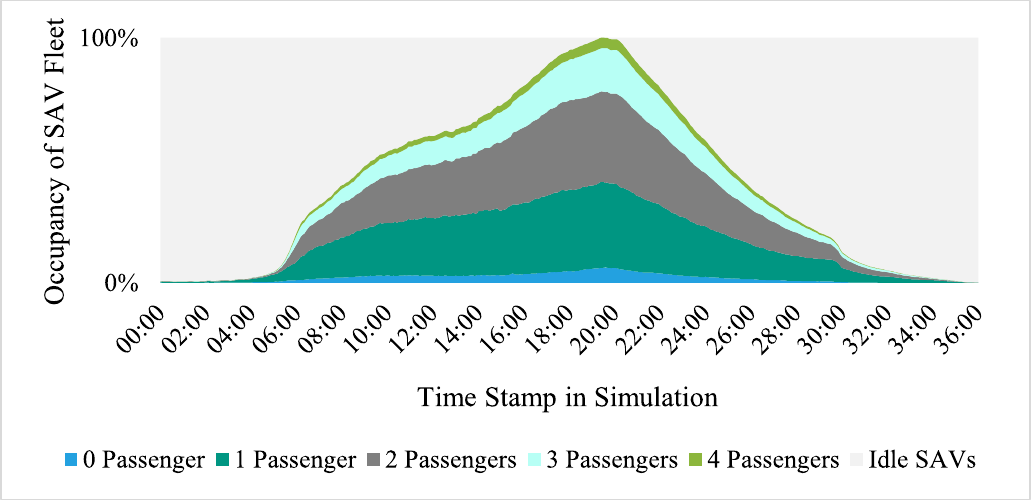}%
    \label{fig:Vehicle_capacity_SC3.1}
  }
  \hfill
  \subfloat[SC3.2.]{%
    \includegraphics[width=0.3\columnwidth]{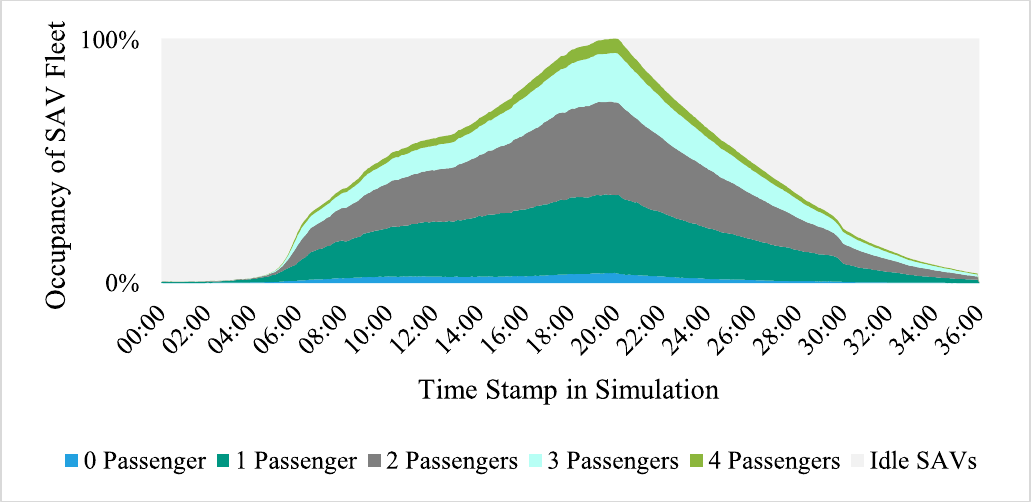}%
    \label{fig:Vehicle_capacity_SC3.2}
  }
  \hfill
  \subfloat[SC3.3.]{%
    \includegraphics[width=0.3\columnwidth]{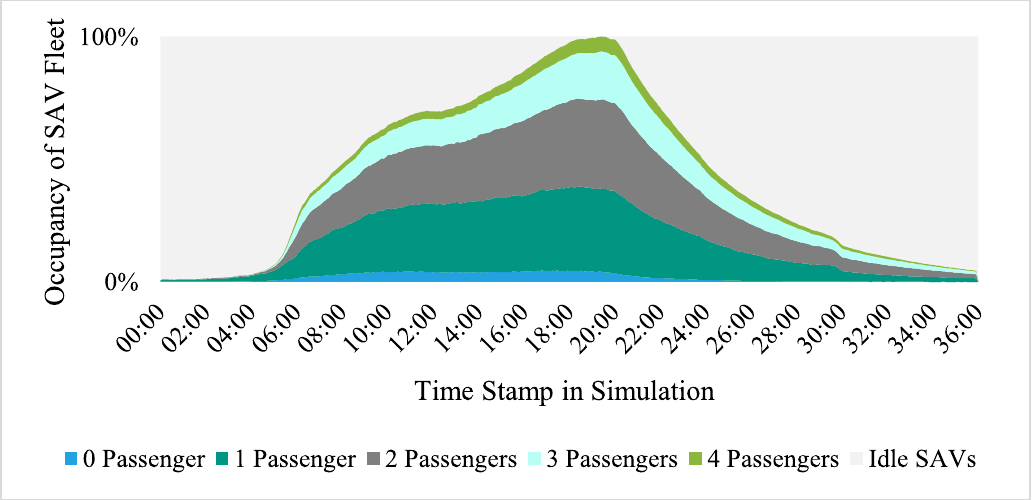}%
    \label{fig:Vehicle_capacity_SC3.3}
  }
  \caption{Vehicle occupancy of SAV fleets. Adapted from~\cite{Karsch_Sustainability_2023_MA}.}
  \label{fig:Vehicle_capacity}
\end{figure}

Especially in the light of sustainability implications, we are interested in the occupation rates of the shared autonomous shuttles. As shown in Figure~\ref{fig:Vehicle_capacity}, the SC3.Y scenarios show comparable patterns. Empty rides are rare, as idle vehicles park in nearby locations. A significant portion of the SAVs carry only one passenger, and also rides with two or three passengers occur frequently. However, a fully occupied SAV with four passengers occurs rarely. For SC1.1-SC1.3, we observe occupation rates of up to 1.67. They increase to up to 1.70 for SC.2.1-SC2.3 and are the highest for SC3.1-SC3.3, ranging from 1.68 to 1.74. These rates are higher than the average occupation rate for private cars in Germany, which is around 1.2 to 1.3~\cite{infas2017mobility}. 

\begin{figure}[h]
  \centering
  \subfloat[100\% fleet size.]{%
    \includegraphics[width=0.23\columnwidth]{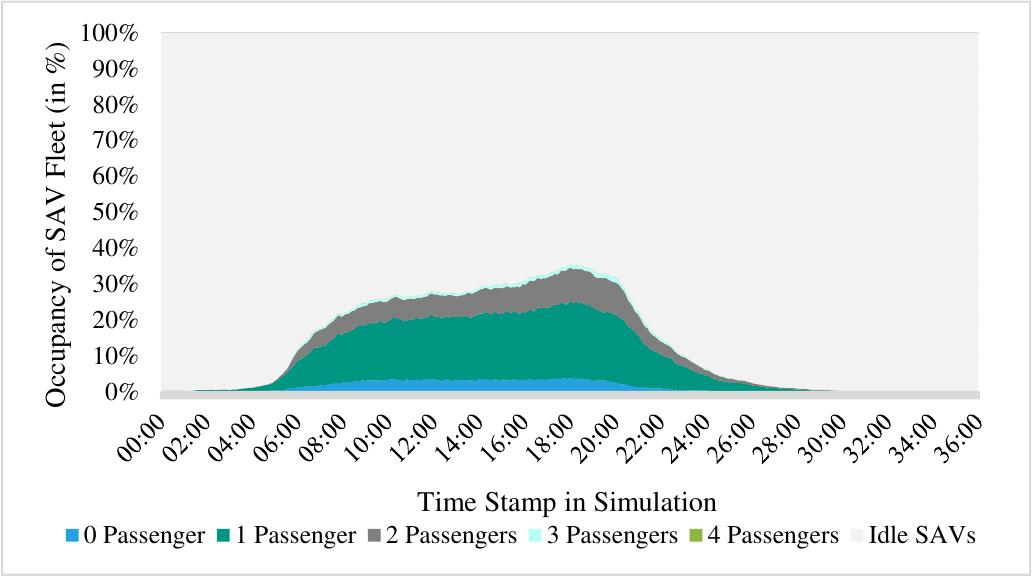}%
    \label{fig:AA_Occupancy_SC1.3}
  }
  \hfill
  \subfloat[60\% fleet size.]{%
    \includegraphics[width=0.23\columnwidth]{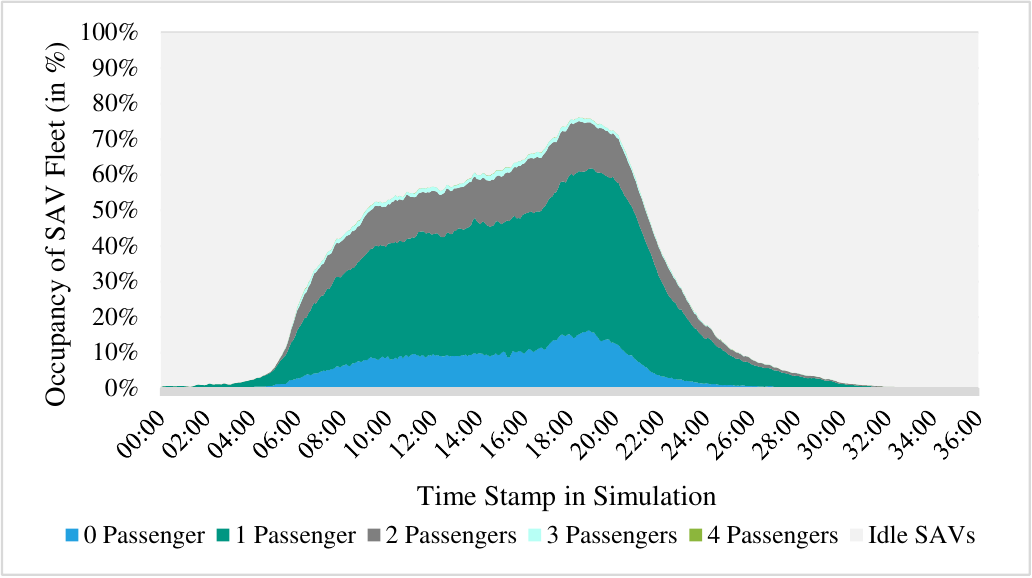}%
    \label{fig:AA_Occupancy_SC1.3_60}
  }
  \hfill
  \subfloat[40\% fleet size.]{%
    \includegraphics[width=0.23\columnwidth]{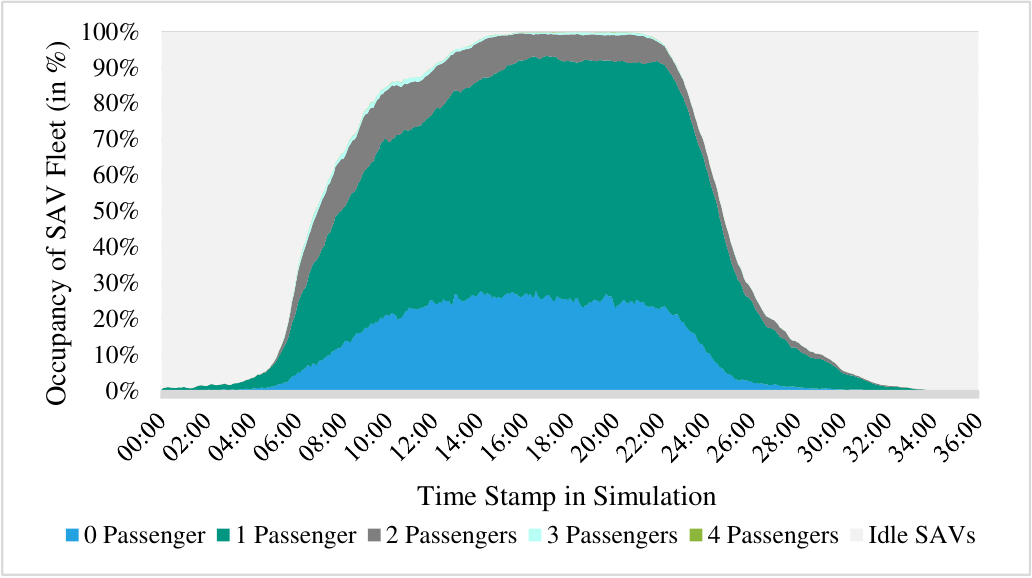}%
    \label{fig:AA_Occupancy_SC1.3_40}
  }
  \hfill
  \subfloat[20\% fleet size.]{%
    \includegraphics[width=0.23\columnwidth]{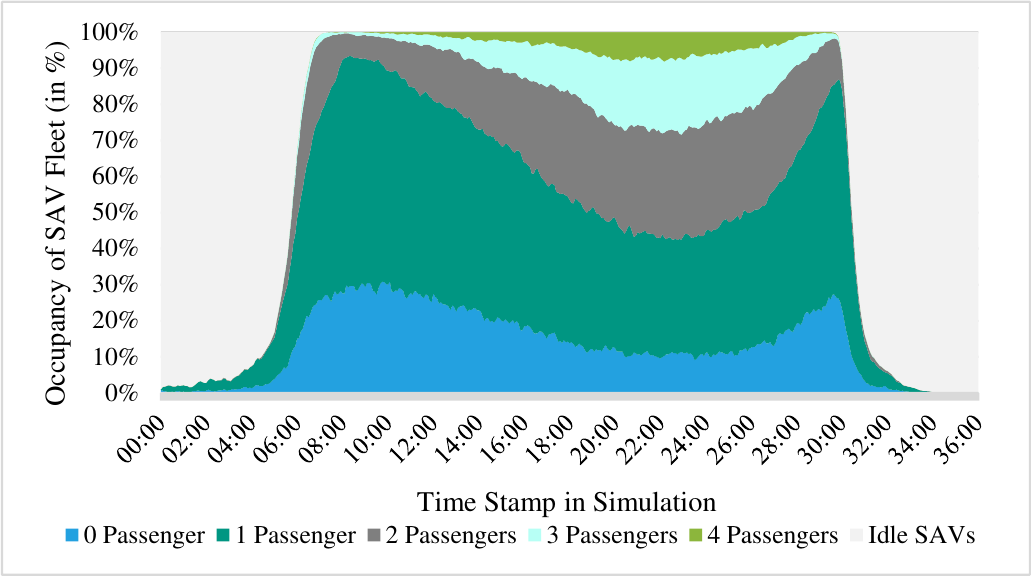}%
    \label{fig:AA_Occupancy_SC1.3_20}
  }
  \caption{Occupancy rates for shared autonomous shuttles for different fleet sizes, on the example of SC1.3. Adapted from~\cite{Karsch_Sustainability_2023_MA}.}
  \label{fig:AA_Vehicle_Occupancy}
\end{figure}

We also examine the effect of artificially reducing fleet sizes for our SC1.3 scenario, neglecting decreased service levels. In our baseline, we allow the simulation to deploy up to 100,000 shuttles. For comparison, we examine scenarios where only 20\% - 60\% of the total shuttles are available, based on the 1\% OpenBerlin scenario. As visible in Figure~\ref{fig:AA_Vehicle_Occupancy}, the baseline scenario SC1.3 differs already in its baseline from the ones shown in Figure~\ref{fig:Vehicle_capacity}, a majority of the rides only carry a single passenger due to the much-reduced travel demand. Reducing the number of available shuttles shows that single passengers travel is continued until no more vehicles are available. Only then will higher occupancy rates occur, as the 20\% scenario shows.

\section{Sustainability Impacts}
\label{sec:sus_impact}

\begin{figure}[tb]
    
    \begin{center}
        \includegraphics[width=1.0\columnwidth,angle=0]{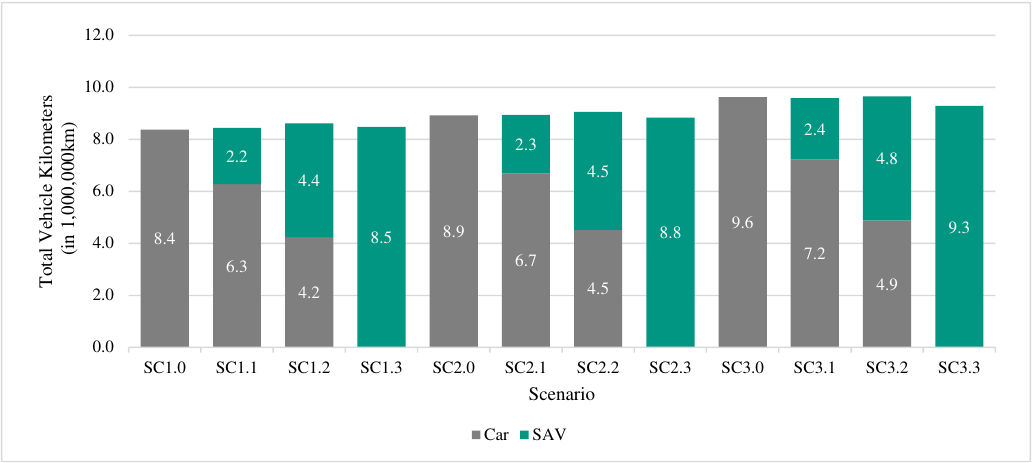}
        \caption{Distance traveled by vehicles. Reprinted from~\cite{Karsch_Sustainability_2023_MA}.}
        \label{fig:Total_Vehicle_Km_Per_Scenario}
    \end{center}
    
\end{figure}

Based on our experiments, we determine the sustainability implications for each scenario. Using total vehicle kilometers, we calculate driving-related emission generation, driving-related energy consumption, and non-driving-related emissions and combine them for a holistic life-cycle assessment, including production, maintenance, and recycling. Figure~\ref{fig:Total_Vehicle_Km_Per_Scenario} shows the traveled distances for cars and SAVs. While we typically first observe an increase in kilometers, increased occupancy rates lead to a relative reduction. For SC2.3, we observe a decrease of up to 1.0\%, which is even higher for SC3.3. going up to 3.6\%. Based on these traveled kilometers, we derive sustainability impacts.

\textbf{Driving-related Emissions.} We consider the influences of powertrain types and energy storage systems separately. Therefore, we distinguish between two perspectives: First, we assume that the SAV fleet will have the same powertrain split as private cars. Second, we assume a fully electric SAV fleet. 

For the first scenario, we analyze the vehicle types present in Berlin and Brandenburg in 2011 and derive the following powertrain split: 78.9\% gasoline, 19.7\% diesel, 0.2\% Hybrid Electric Vehicles (HEV) / Battery Electric Vehicles (BEV) and 1.2\% other vehicles~\cite{kba2011fahrzeugbestand}. 

\begin{table}[h]
\resizebox{\columnwidth}{!}{%
\begin{tabular}{@{}llllllll@{}}
\toprule
                  & \multicolumn{5}{c}{Emissions}                                                                                        & \multicolumn{2}{c}{Energy} \\ \midrule
                  & \multicolumn{2}{c|}{Ours}                            & \multicolumn{3}{c|}{Patella et al.~\cite{patella2019carbon}}                                     & \multicolumn{2}{c}{Ours}               \\ \midrule
                  & \textbf{Urban} & \multicolumn{1}{l|}{\textbf{Rural}} & \textbf{Urban} & \textbf{Rural} & \multicolumn{1}{l|}{\textbf{Average}} & \textbf{Urban}     & \textbf{Rural}    \\
\textbf{Gasoline} & 246.1          & \multicolumn{1}{l|}{146.4}          & 267.91         & 159.38         & \multicolumn{1}{l|}{224.3}            & 0.0282             & 0.0168            \\
\textbf{Diesel}   & 199.4          & \multicolumn{1}{l|}{146.4}          & 216.96         & 159.03         & \multicolumn{1}{l|}{193.7}            & 0.0228             & 0.0168            \\
\textbf{LPG}      & 185.2          & \multicolumn{1}{l|}{157.3}          & 231.38         & 196.49         & \multicolumn{1}{l|}{217.3}            & 0.0280             & 0.0238            \\
\textbf{BEV}      & 0              & \multicolumn{1}{l|}{0}              & 0              & 0              & \multicolumn{1}{l|}{0}                & 0.0084             & 0.0097            \\ \bottomrule
\end{tabular}%
}
\caption{Emissions in g CO$_2$ eq/km per powertrain, as shown on the left, and energy consumption in GGE/km on the right, both for road types of urban and rural areas.}
\label{tab:emissions_energy}
\end{table}

\begin{figure}[htb]
    
    \begin{center}
        \includegraphics[width=\columnwidth,angle=0]{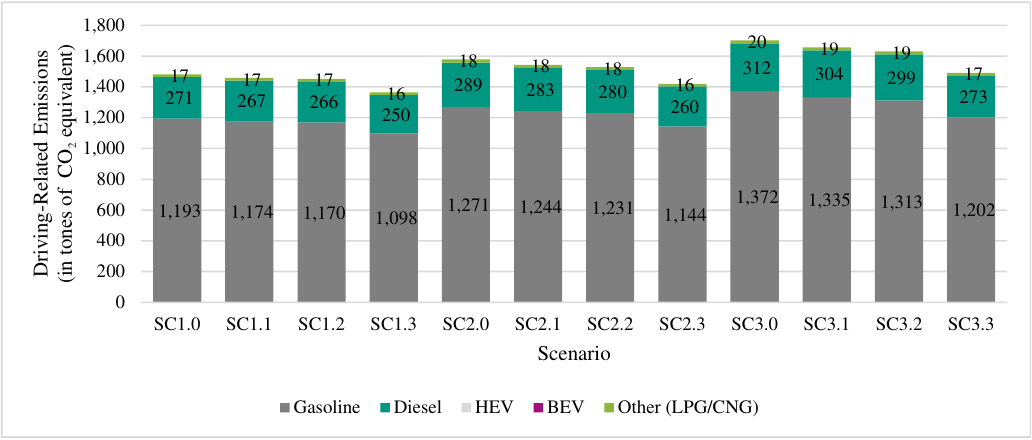}
        \caption{Daily driving-related emissions for an unchanged powertrain split. Reprinted from~\cite{Karsch_Sustainability_2023_MA}.}
        \label{fig:Driving_Related_Emissions}
    \end{center}
    
\end{figure}

Driving-related emissions per powertrain can be found in Table~\ref{tab:emissions_energy}. We base our calculations on the average German emission generation per powertrain from IEA~\cite{iea2019emissions} from 2020 and LPG emissions from EEA/EMEP~\cite{emep2022emissions} from 2021. We assume that by 2050, Berlin and Brandenburg's energy supply will be 100\% based on renewable energy and that BEVs will have 0 grams of CO$_2$ equivalent emissions. For HEVs, we assume urban streets operated electrically and rural streets gasoline-based, as done by Patella et al.~\cite{patella2019carbon}. Streets within the city of Berlin are categorized as urban, while all other streets are classified as rural. We further incorporate efficiency increases of autonomous vehicles~\cite{gawron2018life}. Figure~\ref{fig:Driving_Related_Emissions} shows the emissions per scenario for the unchanged powertrain split. We see decreased values for every scenario that includes SAVs, ranging from 1.6\% to 12.4\%.

\begin{figure}[htb]
    
    \begin{center}
        \includegraphics[width=1.0\columnwidth,angle=0]{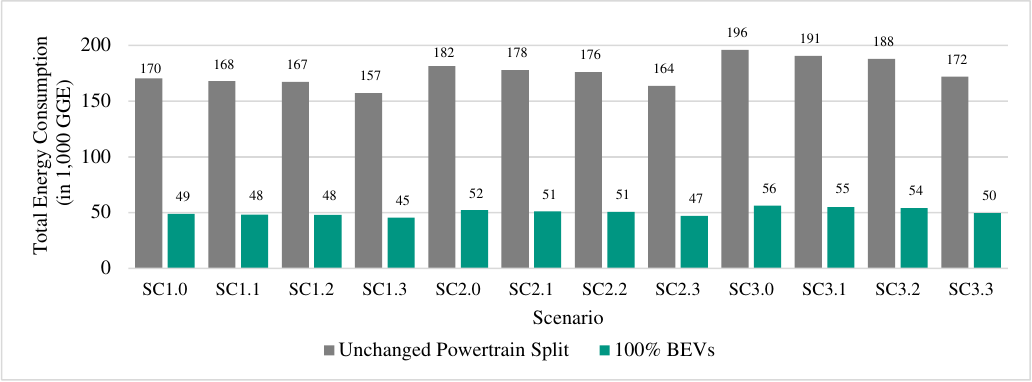}
        \caption{Daily driving-related energy consumption. Reprinted from~\cite{Karsch_Sustainability_2023_MA}}
        \label{fig:Total_Energy_Consumption_2023_2050}
    \end{center}
    
\end{figure}

\textbf{Driving-related Energy Consumption.} Next to emissions, we also examine energy consumption. To improve comparability between scenarios, we use the unit of gasoline gallon equivalents~(GGE), see Table~\ref{tab:emissions_energy}. We further incorporate efficiency increases of autonomous vehicles~\cite{gawron2018life}. Introducing SAVs decreases total energy consumption, as visible in  Figure~\ref{fig:Total_Energy_Consumption_2023_2050}. The largest reduction can be seen for SCX.3, ranging from 7.7\% for SC1.3 to 12.2\% for SC3.3, related to the decrease in total vehicle kilometers. If we combine sharing effects and a fully electric fleet, reductions of up to 71.3\% can be seen.

\begin{figure}[htb]

    \begin{center}
        \includegraphics[width=1.0\columnwidth,angle=0]{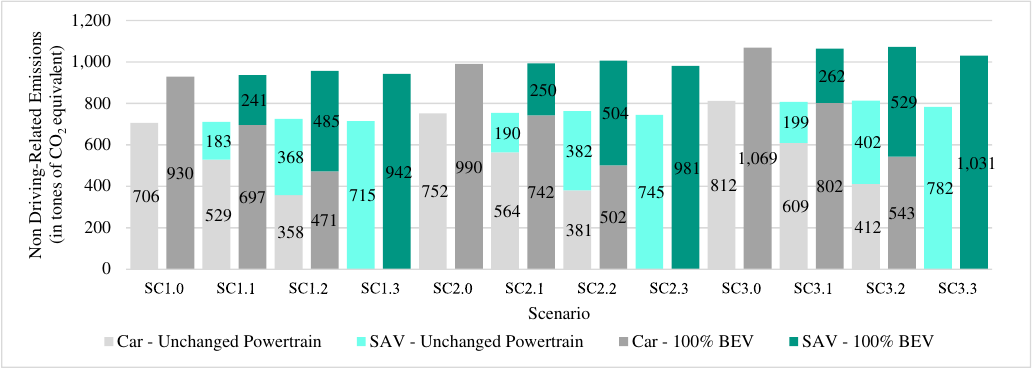}
        \caption{Daily non-driving-related emissions. Reprinted from~\cite{Karsch_Sustainability_2023_MA}.}
        \label{fig:Non_driving_Related_Emissions_2023_vs_2050}
    \end{center}
    
\end{figure}

\textbf{Non-driving-related Emissions.} Here, we derive emissions for construction, maintenance, and end-of-life processes, following Patella et al.~\cite{patella2019carbon}. We can observe that introducing SAVs has a small influence on these emissions. However, non-driving related emissions are significantly higher for fully electric fleets, as shown in Figure~\ref{fig:Non_driving_Related_Emissions_2023_vs_2050}.

\begin{figure}[htb]
    
    \begin{center}
        \includegraphics[width=1.0\columnwidth,angle=0]{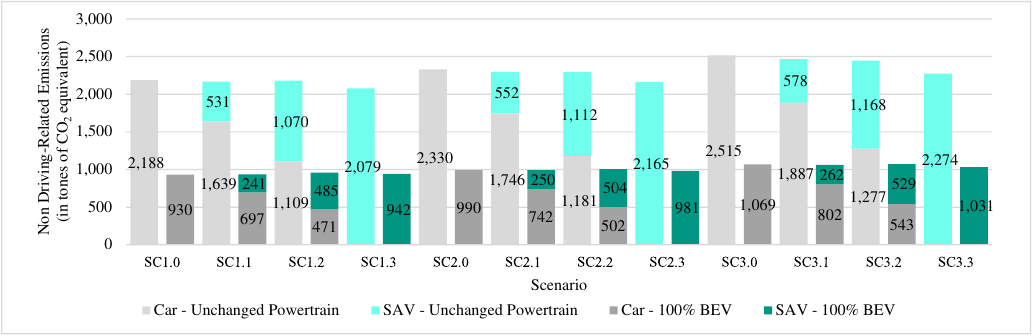}
        \caption{Daily life-cycle emissions. Reprinted from~\cite{Karsch_Sustainability_2023_MA}}
        \label{fig:Total_Daily_Emissions_Car_SAV_2050}
    \end{center}
    
\end{figure}

\textbf{Life-Cycle Assessment.} Finally, we compute emissions for a whole life cycle by combining all three categories. For an unchanged powertrain split, we can observe that driving-related emissions account for up to 67.7\% of total emissions, while the impact of a fully electric fleet is only driven by the non-driving-related emissions, as shown in Figure~\ref{fig:Total_Daily_Emissions_Car_SAV_2050}. It is clear that although the effects of sharing vehicles are positive, they cannot match the scale of the effects of electrification. Accordingly, the highest savings in our scenarios can be achieved by a fully electric, autonomous fleet. 

\section{Evaluation}
\label{sec:evaluation}

In this section, we critically analyze our results, compare them to findings from related works, and highlight limitations.

\textbf{Travel Demand Forecast.} For our forecasts, we focused on age-related travel behavior and population growth, analyzing increased travel demands of up to 14.9\%. However, this does not consider future behavioral shifts introduced by the existence of autonomous vehicles, such as a reduced need for a driver's license and thus increased travel demand~\cite{lutin2013revolutionary,patella2019carbon}. Both younger and older people might show an increased travel demand. Also, due to increased comfort or reduced costs, travel distances might increase compared to today, leading to an increased travel demand of up to 20\%~\cite{sonnleitner2022impacts, massar2021impacts}. Winkler and Mocanu~\cite{winkler2020impact} find vehicle travel demand for Germany increasing up to 6\% until 2040 compared to our 6.6\%. Lutin et al.~\cite{lutin2013revolutionary} assume that AV availability would result in changes in travel demand per age group, potentially increasing our forecasts as well. However, also reduced travel demands are possible in the future, e.g., driven by remote work~\cite{abe2023impact, caros2023evaluating, roberto2023potential}.

\textbf{Agent-Based Simulation.} For our simulation, we integrated the macroscopic forecasts into a microscopic, agent-based simulation environment. As our newly introduced agents are based on agents from the original 2011 Open Berlin Scenario with similar travel patterns, they might lack realism. In addition, we did not incorporate any other changes, such as new traffic infrastructure or increased public transportation capacities. Patella et al.~\cite{patella2019carbon} incorporated capacity increases of 40\% for urban streets and 80\% for highways. The inaccuracies in our simulation might contribute to street capacity bottlenecks, which we could observe in our simulations. We see average waiting time spikes of 20 min for SC1.1 and 10 minutes for SC1.3, which is close to the findings from Bischoff and Maciejewski~\cite{bischoff2016autonomous}, who observed spikes of 14 to 15 min and waiting times of up to 20 min for outer parts of the city.

It is unclear whether our regulatory interventions prohibiting private passenger cars can become a reality. We see both works that have examined similar strategies~\cite{ziemke2023accessibilities, zwick2021agent, bischoff2016autonomous} and such which doubt diminishing vehicle ownership~\cite{pakusch2018unintended,harb2021we}.

As we build our sustainability analysis primarily on kilometers driven, we compare our findings to those of others in the field. In a 100\% AV scenario, Patella et al.~\cite{patella2019carbon} see an increase of 1\% for vehicles distances and found decreases of 5.0\% in urban areas and increases of 8.0\% on highways. They argue that AVs show higher capacity impacts on highways. Their approach is similar to our SC1.3, where we predict an increase of up to 1.3\%. However, their study assumes AVs rather than SAVs. Different pricing strategies were examined by Liu et al.~\cite{liu2017tracking}, where they see shares of shared autonomous vehicles between 9.2\% and 50.9\%. This leads to increased vehicle distances between 9.8\% and 15.7\%. Similar to our findings, they see that vehicle distances don't increase as much anymore when the SAV share is already high. For a 69.4\% SAV share, we see an increased travel demand of 0.8\%, which goes up to 2.8\% for a 38.9\% SAV share.

A scenario where 100\% of travel in Berlin City was performed by autonomous vehicles was provided by Bischoff and Maciejewski~\cite{bischoff2016autonomous}. Our comparable scenarios show that an SAV travels a distance between 203 and 216 km on average, which is similar to their 274 km. Also, trip distances were comparable, with 9.1 km in their experiments and 11.0 km in ours. As we also included traffic leaving and entering the Berlin city area, our number is slightly higher. However, they saw a share of roughly 13\% empty rides, compared to our much lower 3\%. This can be explained by our much larger fleet size, which became necessary due to the including of outer Berlin. In our experiments, up to 68,100 SAVs were deployed for the 10\% Open Berlin Scenario, whereas their experiments see only 100,000 SAVs for the full 100\% Open Berlin Scenario. Also, SAVs stay in close proximity to the agent's home addresses during nighttime.

\textbf{Sustainability Impacts.} Our sustainability analysis includes both driving-related emissions and non-driving related emissions. Neglecting future improvements of ICEVs and differences between warm- and cold-starts~\cite{fagnant2014travel}, we compare a common powertrain split with a fully electrified scenario, as done by Patella et al.~\cite{patella2019carbon}. For this, we assume fully renewable electricity, which affects the calculation of emissions drastically~\cite{huo2015life} and differs strongly from the electricity mix we see today~\cite{umweltbundesamt2021energiemix}. As we include traffic from both urban and rural areas, we also distinguish between these categories when we compute our sustainability metrics, as previously done~\cite{farhan2018impact, patella2019carbon}.

Fagnant et al.~\cite{fagnant2014travel} examined a completely artificial scenario where 100\% of the car-based travel is done by SAVs and found reduced CO$_2$ emissions of 34\% and reduced total emissions of 5.6\%. The analysis of Harper et al.~\cite{harper2018exploring} showed that the introduction of AVs might allow people to travel longer distances as parking becomes less of an issue. This can lead to increases in energy use and greenhouse gas emissions of 3\%. On the other hand, Liu et al.~\cite{liu2017tracking} examined different pricing strategies for the use of SAVs, where their experiments showed a share of SAVs of 50.9\%. They found reduced emissions of 8.7\% and reduced fuel consumption of up to 14.8\%. Comparing these results with our SC1.2, in which SAVs account for 38.9\% of total agent trips, and SC1.3, in which SAVs account for 69.5\% of total agent trips, we determine that there is potential for a reduction in energy consumption ranging from 1.8\% to 7.7\% and a reduction in emissions ranging from 0.4\% to 5.0\%. Simulating a scenario with a fully electrified AV fleet and comparing it to a conventional powertrain split, Patella et al.~\cite{patella2019carbon} see reduced emissions of 59.6\%. While our powertrain split differs slightly, we see can observe similar reductions of 56.9\%.
\section{Conclusion}
\label{sec:conclusion}

We presented an agent-based simulation that outlines the sustainability impacts of three types of SAV introductions in the private passenger transport sector across the larger Berlin area. Our approach builds on the current literature and improves it by including travel demand projections and comprehensive, more holistic sustainability assessments. Based on multiple experiments, our results suggest that introducing shared autonomous vehicles (SAVs) could potentially reduce emissions generation by 0.4-9.6\% and energy consumption by 1.5-12.2\%. These reductions could increase to up to 59.0\% with a 100\% battery electric SAV fleet.

\textbf{Limitations.} Instead of forecasting SAV adoption, this study examines the impact of hypothetical regulations in the form of replacing private passenger cars with shared autonomous shuttles. We do not incorporate changes in travel demand induced by new technologies, e.g., based on low acceptance levels of autonomous vehicles. We only highlight effects on the service level of the SAV fleets, but do not examine them in detail or even optimize for them. However, as we observe high waiting times, there is a need for future improvements to simulate more realistic scenarios. For the agent-based simulations, we did not compare multiple matching strategies for improved ridesharing. Also, as we do not use hubs, our approach requires distributed charging and maintenance. If hubs were in use and placed in a strategic way, empty rides would probably increase. Finally, we build on simple assumptions for a fully renewable future energy supply and do not predict potentially more realistic energy mixes.

\textbf{Outlook.} While we provided two diverse forecasts of travel demands, the granularity of the approach can be further increased, e.g., by incorporating social changes. As we predicted travel demand but no changes in traffic infrastructure, e.g., increased road capacity or changes in public transport, a future Open Berlin Scenario for 2050 would be of interest, also incorporating increased travel demands in earlier stages. Generally, increasing the number of iterations for our simulations would increase the accuracy of our results. Waiting times of agents could be partially improved by a scheduling algorithm~\cite{Bogdoll_DLCSS_2022_ICECCME} on the day ahead, as there is no need for real-time dispatching for regular commutes. An additional mode of privately owned AVs would further improve the diversity of our experiments. Other approaches to SAV adoption, especially more naturally evolving ones, could be compared to our scenarios. Finally, improved calculations for our sustainability impacts are of interest. Here, we would like to incorporate temporal factors that affect energy consumption, such as cold and warm starts, as well as the effects of large-scale remote assistance centers, which are necessary for large SAV fleets~\cite{Bogdoll_Taxonomy_2022_FICC,cruiseremote}. Here, especially the dimensionality of such centers~\cite{gontscharow2023} and their sustainability implications are of interest.

% Im AP3.1 gibt es die Komponente Spezifikation des KI-Anwendungsfalles Lernendes Situationsverstehen für Störungsmanagement und Optimierung des Leitstandbetriebs für Automatisierte Fahrzeuge. Hier wäre mein Konzept, für "Spezifikation/Optimierung des Leitstandbetriebs für Automatisierte Fahrzeuge" zunächst zu beantworten, wie viele Shuttles in einem nachhaltigen Betrieb unterwegs sind. Auf der Basis kann man den Leitstandsbetrieb dimensionieren/spezifizieren/optimieren.

% https://ids-git.fzi.de/project-management/OeV-LeitmotiF-KI/-/issues/7

\section{Acknowledgment}
\label{sec:acknowledgment}

This work results partly from the ÖV-LeitmotiF-KI project supported by the German Federal Ministry for Digital and Transport (BMDV), grant number 45AVF3004C.

% -------------------------- REFERENCES -------------------------------

{\small
\bibliographystyle{IEEEtran}
\bibliography{references}

% Generated by IEEEtran.bst, version: 1.14 (2015/08/26)
\begin{thebibliography}{10}
\providecommand{\url}[1]{#1}
\csname url@samestyle\endcsname
\providecommand{\newblock}{\relax}
\providecommand{\bibinfo}[2]{#2}
\providecommand{\BIBentrySTDinterwordspacing}{\spaceskip=0pt\relax}
\providecommand{\BIBentryALTinterwordstretchfactor}{4}
\providecommand{\BIBentryALTinterwordspacing}{\spaceskip=\fontdimen2\font plus
\BIBentryALTinterwordstretchfactor\fontdimen3\font minus
  \fontdimen4\font\relax}
\providecommand{\BIBforeignlanguage}[2]{{%
\expandafter\ifx\csname l@#1\endcsname\relax
\typeout{** WARNING: IEEEtran.bst: No hyphenation pattern has been}%
\typeout{** loaded for the language `#1'. Using the pattern for}%
\typeout{** the default language instead.}%
\else
\language=\csname l@#1\endcsname
\fi
#2}}
\providecommand{\BIBdecl}{\relax}
\BIBdecl

\bibitem{sudhakar2022data}
S.~Sudhakar \emph{et~al.}, ``Data centers on wheels: emissions from computing
  onboard autonomous vehicles,'' \emph{IEEE Micro}, vol.~43, 2022.

\bibitem{gawron2018life}
J.~H. Gawron \emph{et~al.}, ``Life cycle assessment of connected and automated
  vehicles: sensing and computing subsystem and vehicle level effects,''
  \emph{Environmental science \& technology}, vol.~52, 2018.

\bibitem{massar2021impacts}
M.~Massar \emph{et~al.}, ``Impacts of autonomous vehicles on greenhouse gas
  emissions—positive or negative?'' \emph{International Journal of
  Environmental Research and Public Health}, vol.~18, 2021.

\bibitem{patella2019carbon}
S.~Patella \emph{et~al.}, ``Carbon footprint of autonomous vehicles at the
  urban mobility system level: A traffic simulation-based approach,''
  \emph{Transp. Res. Part D: Transport and Environment}, vol.~74, 2019.

\bibitem{silva2022environmental}
{\'O}.~Silva \emph{et~al.}, ``Environmental impacts of autonomous vehicles: A
  review of the scientific literature,'' \emph{Science of The Total
  Environment}, 2022.

\bibitem{bischoff2016autonomous}
J.~Bischoff and M.~Maciejewski, ``Autonomous taxicabs in berlin--a
  spatiotemporal analysis of service performance,'' \emph{Transp. Res. Proc.},
  vol.~19, 2016.

\bibitem{horl2017agent}
S.~H{\"o}rl, ``Agent-based simulation of autonomous taxi services with dynamic
  demand responses,'' \emph{Proc. Computer Science}, vol. 109, 2017.

\bibitem{liu2017tracking}
J.~Liu \emph{et~al.}, ``Tracking a system of shared autonomous vehicles across
  the austin, texas network using agent-based simulation,''
  \emph{Transportation}, vol.~44, 2017.

\bibitem{zhang2015performance}
W.~Zhang \emph{et~al.}, ``The performance and benefits of a shared autonomous
  vehicles based dynamic ridesharing system: An agent-based simulation
  approach,'' in \emph{Transp. Res. Board 94th Annual Meeting}, vol.~15, 2015.

\bibitem{hidaka2018forecasting}
K.~Hidaka and T.~Shiga, ``Forecasting travel demand for new mobility services
  employing autonomous vehicles,'' \emph{Transp. Res. Proc.}, vol.~34, 2018.

\bibitem{lecureux2021sensitivity}
B.~L{\'e}cureux and I.~Kaddoura, ``Sensitivity of the urban transport system to
  the value of travel time savings for shared autonomous vehicles: A simulation
  study,'' \emph{Proc. Computer Science}, vol. 184, 2021.

\bibitem{vosooghi2019shared}
R.~Vosooghi \emph{et~al.}, ``Shared autonomous vehicle simulation and service
  design,'' \emph{Transp. Res. Part C: Emerging Technologies}, vol. 107, 2019.

\bibitem{ziemke2023accessibilities}
D.~Ziemke and J.~Bischoff, ``Accessibilities by shared autonomous vehicles
  under different regulatory scenarios,'' \emph{Proc. Computer Science}, vol.
  220, 2023.

\bibitem{zwick2021agent}
F.~Zwick \emph{et~al.}, ``Agent-based simulation of city-wide autonomous
  ride-pooling and the impact on traffic noise,'' \emph{Transp. Res. Part D:
  Transport and Environment}, vol.~90, 2021.

\bibitem{bansal2017forecasting}
P.~Bansal and K.~M. Kockelman, ``Forecasting americans’ long-term adoption of
  connected and autonomous vehicle technologies,'' \emph{Transp. Res. Part A:
  Policy and Practice}, vol.~95, 2017.

\bibitem{cokyasar2020analyzing}
T.~Cokyasar \emph{et~al.}, ``Analyzing energy and mobility impacts of
  privately-owned autonomous vehicles,'' in \emph{International Conference on
  Intelligent Transportation Systems (ITSC)}, 2020.

\bibitem{hao2017analysis}
M.~Hao and T.~Yamamoto, ``Analysis on supply and demand of shared autonomous
  vehicles considering household vehicle ownership and shared use,'' in
  \emph{International Conference on Intelligent Transportation Systems (ITSC)},
  2017.

\bibitem{geistfeld2018regulatory}
M.~A. Geistfeld, ``The regulatory sweet spot for autonomous vehicles,''
  \emph{Wake Forest L. Rev.}, vol.~53, 2018.

\bibitem{hansson2020regulatory}
L.~Hansson, ``Regulatory governance in emerging technologies: The case of
  autonomous vehicles in sweden and norway,'' \emph{Research in transportation
  economics}, vol.~83, 2020.

\bibitem{gruvzauskas2018minimizing}
V.~Gru{\v{z}}auskas \emph{et~al.}, ``Minimizing the trade-off between
  sustainability and cost effective performance by using autonomous vehicles,''
  \emph{Journal of Cleaner Production}, vol. 184, 2018.

\bibitem{heard2018sustainability}
B.~R. Heard \emph{et~al.}, ``Sustainability implications of connected and
  autonomous vehicles for the food supply chain,'' \emph{Resources,
  conservation and recycling}, vol. 128, 2018.

\bibitem{williams2020assessing}
E.~Williams \emph{et~al.}, ``Assessing the sustainability implications of
  autonomous vehicles: Recommendations for research community practice,''
  \emph{Sustainability}, vol.~12, 2020.

\bibitem{Karsch_Sustainability_2023_MA}
L.~Karsch, ``{Sustainability of Autonomous Vehicles: An Agent-based Simulation
  of the Private Passenger Sector},'' Master Thesis, {Karlsruhe Institute of
  Technology (KIT)}, 2023.

\bibitem{bridgelall2021forecasting}
R.~Bridgelall and E.~Stubbing, ``Forecasting the effects of autonomous vehicles
  on land use,'' \emph{Technological Forecasting and Social Change}, vol. 163,
  2021.

\bibitem{ben2022modal}
G.~Ben-Dor \emph{et~al.}, ``Modal shift and shared automated demand-responsive
  transport: A case study of jerusalem,'' \emph{Proc. Computer Science}, vol.
  201, 2022.

\bibitem{kaddoura2021impact}
I.~Kaddoura and T.~Schlenther, ``The impact of trip density on the fleet size
  and pooling rate of ride-hailing services: A simulation study,'' \emph{Proc.
  Computer Science}, vol. 184, 2021.

\bibitem{li2023simulation}
J.~Li \emph{et~al.}, ``Simulation of shared autonomous vehicles operations with
  relocation considering external traffic: Case study of brussels,''
  \emph{Proc. Computer Science}, vol. 220, 2023.

\bibitem{harper2018exploring}
C.~D. Harper \emph{et~al.}, ``Exploring the economic, environmental, and travel
  implications of changes in parking choices due to driverless vehicles: An
  agent-based simulation approach,'' \emph{Journal of Urban Planning and
  Development}, vol. 144, 2018.

\bibitem{martinez2017assessing}
L.~M. Martinez and J.~M. Viegas, ``Assessing the impacts of deploying a shared
  self-driving urban mobility system: An agent-based model applied to the city
  of lisbon, portugal,'' \emph{International Journal of Transportation Science
  and Technology}, vol.~6, 2017.

\bibitem{litman2017autonomous}
T.~Litman, ``Autonomous vehicle implementation predictions,'' Victoria
  Transport Policy Institute, Tech. Rep., 2017.

\bibitem{ziemke2019matsim}
D.~Ziemke \emph{et~al.}, ``The matsim open berlin scenario: A multimodal
  agent-based transport simulation scenario based on synthetic demand modeling
  and open data,'' \emph{Proc. computer science}, vol. 151, 2019.

\bibitem{fagnant2014travel}
D.~J. Fagnant and K.~M. Kockelman, ``The travel and environmental implications
  of shared autonomous vehicles, using agent-based model scenarios,''
  \emph{Transp. Res. Part C: Emerging Technologies}, vol.~40, 2014.

\bibitem{cruisetech}
{Bellan, Rebecca}, ``Cruise would join the call to ban human drivers in city
  centers, says ceo,''
  \url{https://techcrunch.com/2023/09/21/cruise-would-join-the-call-to-ban-human-drivers-in-city-centers-says-ceo},
  2023, accessed: 11-11-2023.

\bibitem{genesis2023population}
{Statistisches Bundesamt (Destatis)}, ``Population: L{\"a}nder, reference date,
  sex, age,''
  \url{https://www-genesis.destatis.de/genesis/online?sequenz=statistikTabellen&selectionname=12411#abreadcrumb},
  2023, accessed: 09-07-2023.

\bibitem{infas2017mobility}
{Infas and DLR}, ``Mobilit{\"a}t in deutschland 2018 ergebnisbericht,''
  \emph{BMVBS}, 2017.

\bibitem{w2016multi}
K.~W~Axhausen \emph{et~al.}, \emph{The multi-agent transport simulation
  MATSim}.\hskip 1em plus 0.5em minus 0.4em\relax Ubiquity Press, 2016.

\bibitem{li2021systematic}
J.~Li \emph{et~al.}, ``A systematic review of agent-based models for autonomous
  vehicles in urban mobility and logistics: Possibilities for integrated
  simulation models,'' \emph{Computers, Environment and Urban Systems},
  vol.~89, 2021.

\bibitem{jing2020agent}
P.~Jing \emph{et~al.}, ``Agent-based simulation of autonomous vehicles: A
  systematic literature review,'' \emph{IEEE Access}, vol.~8, 2020.

\bibitem{einheit2015statistische}
{Statistisches Bundesamt (Destatis)}, ``Zensus 2011,''
  \url{https://ergebnisse2011.zensus2022.de/datenbank/online?operation=statistic&code=2000S#abreadcrumb},
  2015, accessed: 27-06-2023.

\bibitem{kba2011fahrzeugbestand}
{Kraftfahrt Bundesamt}, ``Fahrzeugzulassungen (fz) bestand an kraftfahrzeugen
  nach umwelt-merkmalen,''
  \url{https://www.kba.de/DE/Statistik/Produktkatalog/produkte/Fahrzeuge/fz13_b_uebersicht.html},
  2011, accessed: 27-08-2023.

\bibitem{iea2019emissions}
{International Energy Agency}, ``Rated real-world well-to-wheel greenhouse gas
  emissions of new light-duty vehicle sales worldwide by size segment,''
  \url{https://www.iea.org/data-and-statistics/charts}, 2019, accessed:
  28-08-2023.

\bibitem{emep2022emissions}
{European Environment Agency}, ``{EMEP/EEA Air Pollutant Emission Inventory
  Guidebook},'' 2022.

\bibitem{lutin2013revolutionary}
J.~M. Lutin \emph{et~al.}, ``The revolutionary development of self-driving
  vehicles and implications for the transportation engineering profession,''
  \emph{Institute of Transportation Engineers}, vol.~83, 2013.

\bibitem{sonnleitner2022impacts}
J.~Sonnleitner \emph{et~al.}, ``Impacts of highly automated vehicles on travel
  demand: macroscopic modeling methods and some results,''
  \emph{Transportation}, vol.~49, 2022.

\bibitem{winkler2020impact}
C.~Winkler and T.~Mocanu, ``Impact of political measures on passenger and
  freight transport demand in germany,'' \emph{Transp. Res. Part D: Transport
  and Environment}, vol.~87, 2020.

\bibitem{abe2023impact}
R.~Abe \emph{et~al.}, ``Impact of working from home on travel behavior of rail
  and car commuters: A case study in the tokyo metropolitan area,'' \emph{Case
  Studies on Transport Policy}, vol.~11, 2023.

\bibitem{caros2023evaluating}
N.~S. Caros and J.~Zhao, ``Evaluating the travel impacts of a shared mobility
  system for remote workers,'' \emph{Transp. Res. Part D: Transport and
  Environment}, vol. 121, 2023.

\bibitem{roberto2023potential}
R.~Roberto \emph{et~al.}, ``Potential benefits of remote working on urban
  mobility and related environmental impacts: Results from a case study in
  italy,'' \emph{Applied Sciences}, vol.~13, 2023.

\bibitem{pakusch2018unintended}
C.~Pakusch \emph{et~al.}, ``Unintended effects of autonomous driving: A study
  on mobility preferences in the future,'' \emph{Sustainability}, vol.~10,
  2018.

\bibitem{harb2021we}
M.~Harb \emph{et~al.}, ``What do we (not) know about our future with automated
  vehicles?'' \emph{Transp. Res. part C: emerging technologies}, vol. 123,
  2021.

\bibitem{huo2015life}
H.~Huo \emph{et~al.}, ``Life-cycle assessment of greenhouse gas and air
  emissions of electric vehicles: A comparison between china and the us,''
  \emph{Atmospheric Environment}, vol. 108, 2015.

\bibitem{umweltbundesamt2021energiemix}
{Umweltbundesamt}, ``Erneuerbare energien in deutschland daten zur entwicklung
  im jahr 2021,'' Umweltbundesamt, Tech. Rep., 2021.

\bibitem{farhan2018impact}
J.~Farhan and T.~D. Chen, ``Impact of ridesharing on operational efficiency of
  shared autonomous electric vehicle fleet,'' in \emph{Transp. Res. Part C:
  Emerging Technologies}, 2018.

\bibitem{Bogdoll_DLCSS_2022_ICECCME}
D.~Bogdoll \emph{et~al.}, ``{DLCSS: Dynamic Longest Common Subsequences},'' in
  \emph{IEEE International Conference on Electrical, Computer, Communications
  and Mechatronics Engineering (ICECCME)}, 2022.

\bibitem{Bogdoll_Taxonomy_2022_FICC}
------, ``{Taxonomy and Survey on Remote Human Input Systems for Driving
  Automation Systems},'' in \emph{Future of Information and Communication
  Conference (FICC)}, 2022.

\bibitem{cruiseremote}
{CNBC}, ``Cruise confirms robotaxis rely on human assistance every four to five
  miles,''
  \url{https://www.cnbc.com/2023/11/06/cruise-confirms-robotaxis-rely-on-human-assistance-every-4-to-5-miles.html},
  2023, accessed: 11-11-2023.

\bibitem{gontscharow2023}
M.~Gontscharow \emph{et~al.}, ``Survey and design concept for control centers
  of automated vehicle fleets,'' in \emph{IEEE International Conference on
  Intelligent Transportation Systems (ITSC)}, 2023.

\end{thebibliography}
}

\end{document}